# A Common XML-based Framework for Syntactic Annotation

Nancy Ide†, Laurent Romary‡ and Tomaž Erjavec*

†Department of Computer Science
Vassar College
Poughkeepsie, New York 12604-0520, USA
`ide@cs.vassar.edu`

‡ LORIA/CNRS
Campus Scientifique, B.P. 239
54506 Vandoeuvre-lès-Nancy, FRANCE
`romary@loria.fr`

* Department of Intelligent Systems
Institute "Jožef Stefan"
Jamova 39,
SI-1000 Ljubljana, SLOVENIA
`tomaz.erjavec@ijs.si`

## Abstract

It is widely recognized that the proliferation of annotation schemes runs counter to the need to re-use language resources, and that standards for linguistic annotation are becoming increasingly mandatory. To answer this need, we have developed a framework comprised of an abstract model for a variety of different annotation types (e.g., morpho-syntactic tagging, syntactic annotation, co-reference annotation, etc.), which can be instantiated in different ways depending on the annotator's approach and goals. In this paper we provide an overview of the framework, demonstrate its applicability to syntactic annotation, and show how it can contribute to comparative evaluation of parser output and diverse syntactic annotation schemes.

## 1    Introduction

It is widely recognized that the proliferation of annotation schemes runs counter to the need to re-use language resources, and that standards for linguistic annotation are becoming increasingly mandatory. In particular, there is a need for a general framework for linguistic annotation that is flexible and extensible enough to accommodate different annotation types and different theoretical and practical approaches, while at the same time enabling their representation in a "pivot" format that can serve as the basis for comparative evaluation of parser output, such as PARSEVAL (Harrison, *et al.,* 1991), as well as the development of reusable editing and processing tools.

To answer this need, we have developed such a framework comprised of an abstract model for a variety of different annotation types (e.g., morpho-syntactic tagging, syntactic annotation, co-reference annotation, etc.), which can be instantiated in different ways depending on the annotator's approach and goals. We have implemented both the abstract model and various instantiations using XML schemas (Thompson, *et al.,* 2000), the Resource Definition Framework (RDF) (Lassila and Swick, 2000) and RDF schemas (Brickley and Guha, 2000), which enable description and definition of abstract data models together with means to interpret, via the model, information encoded according to different conventions. The results have been incorporated into XCES (Ide, *et al.,* 2000a), part of the EAGLES Guidelines developed by the Expert Advisory Group on Language Engineering Standards

(EAGLES), [1] which provides a ready-made, standard encoding format together with a data architecture designed specifically for linguistically annotated corpora.

In this paper we provide an overview of our encoding framework and demonstrate its applicability to syntactic annotation. The framework has also been applied to terminology (Terminological Markup Framework[2], ISO project n.16642) and computational lexicons (Ide, *et al.*, 2000b), thus demonstrating its general applicability for a variety of linguistic annotation types. We also show how the framework can contribute to comparative evaluation of diverse syntactic annotation schemes and enable assessment of the relative strengths and weaknesses of different parsers, an area of increasing importance for the field.

## 2    Current practice

Syntactic annotation schemes provide at least one, and often both, of the following kinds of information:

(1) *category* : bracketing and labeling of strings based on syntactic category (e.g., noun phrase, prepositional phrase);

(2) *dependency*:    relations    among    bracketed elements reflecting their syntactic roles (e.g., subject, object).

For example, the annotation in Figure 1, drawn from the Penn Treebank II[3] (hereafter, PTB), uses LISP-like list structures to bracket strings and provide category information. Hierarchical nesting and left-to-right adjacency are used to reflect dependency relations; however, "subject" is explicitly (and redundantly, given that the relation is implicit in the structure) labeled along with the category information, although other relations (e.g., "object") are left implicit in the structure. Relations among non-contiguous elements demand a special numbering mechanism to enable reference to a non-contiguous elements, which enables specifying the NP-SBJ of the embedded sentence by reference to the earlier NP-SBJ-1 node.

```
((S          (NP-SBJ-1 Jones)
   (VP followed)
   (NP him)
   (PP-DIR into
      (NP the front room))
   ,
   (S-ADV (NP-SBJ *-1)
      (VP  closing
          (NP the door)
          (PP    behind
              (NP him)))))
 .))
```

Figure 1 : PTB annotation example

Although they differ in the labels and in some cases the function of various nodes in the tree, most syntactic annotation schemes rely on hierarchical nesting to indicate component relations among elements, together with linear, left-to-right ordering at any level of the tree to indicate grammatical role. In the PTB example, the nesting of the NP "*the front room*" implies that the NP is the object of the prepositional phrase, whereas the position of the NP "*him*" following and at the same level as the VP node implies that this NP is the grammatical object. When it is necessary to specify a relation among non-contiguous elements, any of a variety of *ad hoc* mechanisms may be employed (such as the node numbering scheme in the PTB example). In contrast, pure dependency annotation schemes, such as the scheme described in Carroll, Minnen and Briscoe (forthcoming), specify grammatical relations among elements explicitly; for example, the sentence "*Paul intends to leave IBM*" could be represented as shown in Figure 2, where the predicate is the relation type, the first argument is the head, the second the dependent, and additional arguments may provide category-specific information (e.g., *introducer* for prepositional phrases, etc.). Although dependency schemes do not rely on hierarchical nesting to indicate relations, an independently specified relation hierarchy is typically defined which enables construction of a syntax tree from the dependency annotation.

```
subj(intend,Paul,_)
xcomp(intend,leave,to)
subj(leave,Paul)
dobj(leave,IBM,_)
```

Figure 2 : Dependency annotation according to Carroll, Minnen, and Briscoe

## 3    A model for syntactic annotation

The goal in the XCES is to provide a framework for annotation that is theory and tagset independent. We accomplish this by treating the description of any specific syntactic annotation scheme as a process involving several knowledge sources that interact at various levels. The process allows one to specify, on the one hand, the informational properties of the scheme (i.e., its capacity to represent a given piece of information), and, on the other, the way the scheme can be instantiated (e.g., as an XML document).

Figure 3 shows the overall architecture of the XCES framework for syntactic annotation.

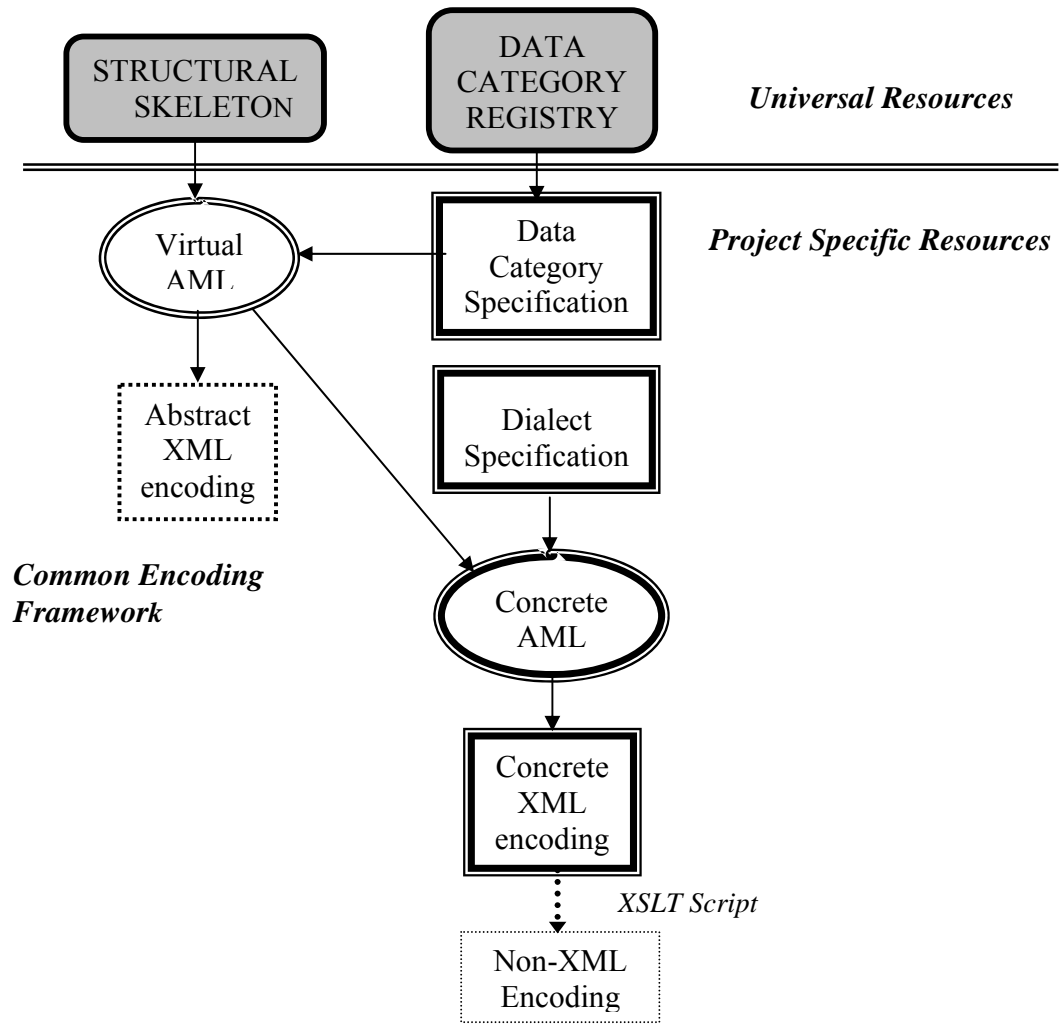

Figure 3 : Overall architecture of the XCES annotation framework

Two universal knowledge sources are used define the abstract model:

**Structural Skeleton**: a domain-dependent abstract structural framework for syntactic annotations, capable of capturing all the information in a specific annotation scheme. The structural skeleton for syntactic annotations is described below in section 3.1.

**Data Category Registry**: Within the framework of the XCES we are establishing an inventory of data categories for syntactic annotation, initially based on the EAGLES

Recommendations for Syntactic Annotation of Corpora (Leech *et al.*, 1996). Data categories are defined using RDF schema that formalize the properties associated with each. The categories are organized in a hierarchy, from general to specific. For example, a general *dependent* relation may be defined, which may be defined to have one of the possible values *argument* or *modifier*; *argument* in turn may have the possible values *subject, object,* or *complement*; etc.[4] Note that RDF schemas function much like class definitions in an object-oriented programming language: they provide, effectively, templates that describe how objects may be instantiated, but do not constitute the objects themselves. Thus, in a document containing an actual annotation, several objects with the type *argument* may be instantiated, each with a different value. The RDF schema ensures recognition that each instantiation of *argument* is a sub-class of *dependent* and inherits the appropriate properties.

The project-specific format for the annotation scheme, in terms of its expressive power and its instantiation in XML is defined in the two other knowledge sources:

**Data Category Specification** (DCS): describes the set of data categories that can be used within a given annotation scheme, again using RDF schema. The DCS defines constraints on each category, including restrictions on the values they can take (e.g., "text with markup"; a "picklist" for grammatical gender, or any of the data types defined for XML), restrictions on where a particular data category can appear (level in the structural hierarchy). The DCS may include a subset of categories from the DCR together with application-specific categories additionally defined in the DCS. The DCS also indicates a level of granularity based on the DCR hierarchy.

**Dialect specification**: defines, using XML schemas, XSLT scripts, and XSL style sheets, the project-specific XML format and labels for syntactic annotations. The specifications may include:

*Data category instantiation styles*: Data categories may be realized in a project-specific scheme in any of a variety of formats. For example, if *NounPhrase* is the value of a complex data category *Cat, Cat* may be realized as an element (`<cat>nom</cat>`), an attribute (`cat="NP"` attached to an anchoring element), a typed element (e.g., `<gram type="cat">NP</gram>`), a valued element (`<cat value="NP"/>`, etc.

*Data category vocabulary styles*: Project-specific formats can utilize names different from those in the Data Category Registry; for instance, a DCR specification for *NounPhrase* can be expressed as "NP" or "SN" ("syntagme nominal", in French) in the project-specific format, if desired.

*Expansion structures*: A project-specific format may alter the structure of the annotation as expressed using the structural skeleton. For example, it may be desirable for processing or other reasons to create two sub-nodes under a given `<struct>` node, one to group features and one to group relations (see section 3.1).

The combination of the structural skeleton and the DCS defines a **virtual annotation markup language (AML)**. Any information structure that corresponds to a virtual AML has a canonical expression as an XML document; therefore, the inter-operability of different AMLs is dependent only on their compatibility at the virtual level. As such, virtual AML is the hub of the annotation framework: it defines the pivot format for syntactic annotations that can be used to compare annotation schemes as well as enable the design of generic tools for visualization, editing, extraction, etc.

The combination of a virtual AML with the Dialect specification provides the information necessary to automatically generate a **concrete AML** representation of the annotation scheme, which conforms to the project-specific specification provide in the Dialect specification. XSLT filters translate between the representations of the annotation in concrete and virtual AML, as well as between non-XML formats (such as the LISP-like PTB notation) and concrete AML.

### 3.1 The structural skeleton

For syntactic annotation, we can identify a general, underlying model that informs current practice: specification of dependency relations (with some set of application-specific names and properties) among a set of grammatical constituents (also with a set of application-specific names and properties), whether this is modeled with a tree structure or the relations are given explicitly.

---

[4] Cf. the hierarchy in Figure 1.1, Caroll, Minnen, and Briscoe (forthcoming).

The most common representation for syntactic annotation in both treebanks and parser output based on manually-developed generative grammars is the phrase-structure tree. Because documents marked in XML are also hierarchically-structured, we provide a structural skeleton for syntactic markup that follows this approach. The skeleton consists of the following elements:

`<struct>` represents a node (level) in the tree. `<struct>` elements may be recursively nested at any level to reflect the structure of the corresponding syntax tree. In the virtual AML, `<struct>` elements are not typed (i.e., associated via attributes with specific data categories such as *Sentence, NounPhrase,* etc.), although project-specific XML schema can provide this information.

`<feat>` (feature) is used to provide information attached to the node in the tree represented by the enclosing `<struct>` element. A *type* attribute on the `<feat>` element identifies the data category of the feature. The tag may contain a string that provides an appropriate value for the data category (e.g., for *type=CAT* the value might be "NP") or may point via a *target* attribute to an object in another document that provides the value. Note that this allows the possibility for including not only simple values but also complex data items as annotations. It also enables generating a single instantiation of an annotation value in a separate document that can be referenced as needed within the annotation document.

`<brack>` is a general-purpose bracketing element to group associated features; it can be recursively nested to describe complex structures.

`<alt>` provides for encoding one or more alternative annotations (e.g., attachment of a PP to the verb and to the preceding object, where ambiguous or undetermined). Together, `<feat>`, `<brack>`, and `<alt>` can be used to encode simple or complex feature structures.

`<rel>` is used to identify dependencies explicitly, by pointing to a `<struct>` element to which the current node is related. The `<rel>` element has the attributes *type, head, dependent, introducer,* which reference the relevant strings or structures, and *initial,* which specifies a thematic or semantic role.

`<seg>` points to the data to which the annotation applies. In XCES, we recommend the use of *stand-off annotation,* i.e., annotation that is maintained in a document separate from the primary (annotated) data. The `<seg>` element would use the *xlink* attribute, which would use the XML Path Language, XPath (Clark & DeRose, 1999) to specify the location of the relevant data.

Figure 4 shows the annotation from the PTB example (Figure 1) rendered in the abstract XML format. Our encoding makes explicit relations that were implicit in the original version; however, implicit relations could be made explicit via an XML schema rather than encoding them explicitly. A strict dependency annotation can be encoded using a flat hierarchy (i.e., a single enclosing `<struct>` element) and specifying all relations explicitly with the `<rel>` tag, as shown in Figure 5. For the sake of readability, this encoding assumes that the sentence is marked up as:

```
<s id= "s1">
<w id="w1">Paul</w>
<w id="w2">intends</w>
<w id="w3">to</w>
<w id ="w4">leave</w>
<w id="w5">IBM</w>
</s>.
```

Note that we encode explicitly here much information that will ultimately be specified using RDF, which provides built-in support for linkage with labeled relations (links). In particular, information currently specified using the `<feat>` element can be represented as a 3-tuple (*resource, property, value*) that indicates that the resource (identified by ID=s1 in the encoding above) has the property CAT with value NP, etc. Similarly, information represented in the encoding above using `<rel>` elements and the *ref* attribute could also be expressed in RDF. Because universal web-based support for RDF is still under development, we provide alternative XML formats for the interim.

```
<struct xml:base="http://www.loria.fr/doc.xml#">
    <struct id="s0">
        <feat type="CAT">S</feat>
        <struct id="s1">
            <feat type="CAT">NP</feat>
            <rel type="SBJ" head="s2"/>
            <seg xlink:href="xptr(substring(/p/s[1]/text(),1,5))"/></struct>
        <struct id="s2">
            <feat type="CAT">VP</feat>
            <seg xlink:href="xptr(substring(/p/s[1]/text(),7,8))"/>
            <struct>
                <feat type="CAT">NP</feat>
                <rel type="OBJ" head="s2"/> <!-- implicit in original -->
                <seg xlink:href="xptr(substring(/p/s[1]/text(),16,3))"/></struct>
        <struct>
            <feat type="CAT">PP</feat>
            <rel type="DIR" head="s2"/>
            <seg xlink:href="xptr(substring(/p/s[1]/text(),20,4))"/>
            <struct>
                <feat type="CAT">NP</feat>
                <seg xlink:href="xptr(substring(/p/s[1]/text(),25,14))"/></struct>
        <struct>
            <feat type="CAT">S</feat>
            <rel type="ADV" head="s2"/>
            <struct ref="s1">
                <feat type="CAT">NP</feat>
                <rel type="SBJ" head="s3"/></struct>
            <struct id="s3">
                <feat type="CAT">VP</feat><struct>…
```

Figure 4 : The PTB example encoded according to the structural skeleton

```
<struct>
    <rel type="subj"  head="w2" dependent="w1"/>
    <rel type="xcomp" head="w2" dependent="w4"  introducer="w3"/>
    <rel type="subj"  head="w4" dependent="w1"/>
    <rel type="dobj"  head="w4" dependent="w5"/>
</struct>
```

Figure 5 : Abstract XML encoding for the dependency annotation in Figure 2.

## 4      Comparing schemes

The Virtual AML provides a pivot format that enables direct comparison of annotations in different formats—including not only different constituency-based annotations, but also constituency-based and dependency annotations. For example, the PTB annotation corresponding to the dependency annotation in Figure 2 is shown in Figure 6. Figure 7 gives the corresponding encoding in the XCES abstract scheme. Relations are encoded only when they appear explicitly in the original annotation; an XML schema specifies the relations implicit in the embedding (e.g., indicating that the first VP is the head, etc.)

From the encoding in Figure 7, an XSLT script[5] produces a dependency encoding virtually identical to the one in Figure 5, thus enabling direct comparison of the two annotations. The only differences between the two would be in relation labels (e.g., XCOMP), which would be mapped via the Dialect Specification to the same category in the Data Category Registry and therefore recognized as equivalent.

The ability to generate a common representation for different annotations overcomes several obstacles that have hindered evaluation exercises in the past. For instance, the evaluation technique used in the PARSEVAL exercise is applicable to phrase-structure analyses only, and cannot be applied to dependency-style analyses or "lexical" parsing frameworks such as finite-state constraint parsers. As the example above shows, this problem is eliminated using the XCES framework.

---

[5] The XML schema and XSLT script are not reproduced here due to space constraints.

```
((S  (NP-SBJ-1 Paul)
            (VP intends)
            (S    (NP-SBJ *-1)
                  (VP  to
                 (VP  leave
                              (NP IBM))))
  .))
```

Figure 6: PTB annotation of "Paul intends to leave IBM."

It has also been noted that that the PARSEVAL bracket-precision measure penalizes parsers that return more structure than exists in the relatively "flat" treebank structures, even if they are correct (Srinivas, *et al.*, 1995). XSLT scripts can extract the appropriate information for comparison purposes while retaining links to additional parts of the annotation in the original document, thus eliminating the need to "dumb down" parser output in order to participate in the evaluation exercise. Similarly, information lost in the transduction from phrase-structure to a dependency-based analysis (as in the example above), which, as Atwell (1996) points out, may eliminate grammatical information potentially required for later processing, can also be retained.

Of course, one of the greatest obstacles to the comparison and evaluation of syntactic annotations remains the difficulty of mapping schemes that are often based on different theories and developed for very different purposes. At the least, our attempt to develop an abstract model that captures the fundamental properties of syntactic annotation schemes will serve, we hope, as a conceptual tool for assessing the coherence and consistency of existing schemes as well as newly developed ones, which is an essential step toward harmonization. The requirement that all relations are made explicit, either in the encoding or via an XML schema that describes them, can lead to the discovery of gaps and inconsistencies in existing schemes. [6] Hand-developed annotation schemes used in treebanks are often described only informally in guidebooks for annotators, often resulting in virtual anarchy; for example, Charniak (1996) notes that the PTB implicitly contains more than 10,000 context-free rules, most of which are used only once. Mappability of schemes becomes virtually impossible under such circumstances. While requiring that annotators make relations explicit and consider the mapping to the XCES abstract format increases overhead, we feel that the exercise will help avoid such problems and can only lead to greater coherence, consistency, and inter-operability among annotation schemes.

The most important contribution to inter-operability of annotation schemes is the Data Category Registry. By mapping site-specific categories onto definitions in the Registry, equivalences (and non-equivalences) are made explicit. Again, the provision of a "standard" set of categories, together with the requirement that scheme-specific categories are mapped to them where possible, can only contribute to greater consistency and commonality for annotation schemes.

# 5    Conclusion

The framework presented here for syntactic annotation is intended to allow for variation in annotation schemes while at the same time enabling comparison and evaluation, as well as the development of common tools for creating and using annotated data. Despite its seeming complexity, the framework is also intended to reduce overhead for annotators and users. Part of the work of the XCES is to provide XML support (e.g., development of XSLT scripts, XML schemas, etc.) for use by the research community, thus eliminating the need for XML expertise at each development site. Also, with XML at its core, the XCES framework builds on emerging standards for data interchange for which there is widespread support; common scripts and tools can be developed, reducing the "reinventing of the wheel" that is commonplace in the area of corpus annotation at present. Because XML-encoded annotated corpora are increasingly used for interchange between individual processing and

---

[6] Indeed, several annotation scheme developers have been surprised to see inconsistencies revealed when

their formats are rendered in the abstract format.

analytic tools, for commonly used tools we are developing XSLT scripts for mapping, and extraction of annotated data, import/export of (partially) annotated material, and integration of results of external tools into existing annotated data in XML. Tools for editing annotations in the abstract format and for automatically generating virtual AML from Data Category and Dialect Specifications are already under development in the context of work on the Terminological Markup Language; as XML use becomes more widespread, more and more reusable tools and resources (including web-based applications) will become available.

```
<struct id="s0">
        <feat type="CAT">S</feat>
        <struct id="s1">
                <feat type="CAT">NP</feat>
                <rel type="SBJ" head="s2"/>
                <seg target="w1"/></struct>
        <struct id="s2">
                <feat type="CAT">VP</feat>
                <seg target="w2"/></struct>
        <struct id="s3">
                <feat type="CAT">S</feat>
                <struct id="s4" ref="s1">
                        <rel type="SBJ" head="s6"/></struct>
                <struct id="s5">
                        <feat type="CAT">VP</feat>
                        <seg target="w3"/>
                        <struct id="s6">
                                <feat type="CAT">VP</feat>
                                <seg target="w4"/>
                                <struct" id="s7">
                                        <feat type="CAT">NP</feat>
                                        <seg target="w5"/></struct>
</struct></struct></struct>
```

Figure 7 : PTB encoding of "*Paul intends to leave IBM.*"